\documentclass{article}

\usepackage{PRIMEarxiv}

\usepackage[utf8]{inputenc} 
\usepackage[T1]{fontenc}    
\usepackage{hyperref}       
\usepackage{url}            
\usepackage{booktabs}       
\usepackage{amsfonts}       
\usepackage{nicefrac}       
\usepackage{microtype}      
\usepackage{lipsum}
\usepackage{fancyhdr}       
\usepackage{graphicx}       

\usepackage{xcolor}
\usepackage{todonotes}
\usepackage{amsmath}
\usepackage{subcaption}

\usepackage{amsfonts}

\newcommand{\model}{ALAG }
\newcommand{\modelns}{ALAG}
\newcommand{\longmodel}{Automated Long Answer Grading }

\newcommand{\dataset}{RiceChem }
\newcommand{\datasetns}{RiceChem}

\usepackage{listings}
\usepackage{xcolor}       
\usepackage{multirow}
\usepackage{caption} 
\colorlet{lightyellow}{yellow!40}

\makeatletter
{\small 
\xdef\f@size@small{\f@size}
\xdef\f@baselineskip@small{\f@baselineskip}
\normalsize 
\xdef\f@size@normalsize{\f@size}
\xdef\f@baselineskip@normalsize{\f@baselineskip}
}
\newcommand{\smalltonormalsize}{%
  \fontsize
    {\fpeval{(\f@size@small+\f@size@normalsize)/2}}
    {\fpeval{(\f@baselineskip@small+\f@baselineskip@normalsize)/2}}%
  \selectfont
}
\makeatother

\title{Automated Long Answer Grading \\ with RiceChem Dataset}

\author{
  Shashank Sonkar\thanks{Equal contribution.}\\
  Rice University \\
  Houston, TX \\
  \texttt{ss164@rice.edu} \\
  \And
  Kangqi Ni\footnotemark[1]\\
  Rice University \\
  Houston, TX \\
  \texttt{kn22@rice.edu} \\
  \And
  Lesa Tran Lu \\
  Rice University \\
  Houston, TX \\
  \texttt{lesa@rice.edu} \\
  \And
  Kristi Kincaid  \\
  Rice University \\
  Houston, TX \\
  \texttt{kk27@rice.edu} \\
  \AND
  John S. Hutchinson \\
  Rice University \\
  Houston, TX \\
  \texttt{jshutch@rice.edu} \\
  \And
  Richard G. Baraniuk \\
  Rice University \\
  Houston, TX \\
  \texttt{richb@rice.edu} \\
}

\begin{document}
\maketitle

\begin{abstract}
We introduce a new area of study in the field of educational Natural Language Processing: Automated Long Answer Grading (ALAG). Distinguishing itself from Automated Short Answer Grading (ASAG) and Automated Essay Grading (AEG), ALAG presents unique challenges due to the complexity and multifaceted nature of fact-based long answers. To study ALAG, we introduce RiceChem, a dataset derived from a college chemistry course, featuring real student responses to long-answer questions with an average word count notably higher than typical ASAG datasets. We propose a novel approach to ALAG by formulating it as a rubric entailment problem, employing natural language inference models to verify whether each criterion, represented by a rubric item, is addressed in the student's response. This formulation enables the effective use of MNLI for transfer learning, significantly improving the performance of models on the RiceChem dataset. We demonstrate the importance of rubric-based formulation in ALAG, showcasing its superiority over traditional score-based approaches in capturing the nuances of student responses. We also investigate the performance of models in cold start scenarios, providing valuable insights into the practical deployment considerations in educational settings. Lastly, we benchmark state-of-the-art open-sourced Large Language Models (LLMs) on RiceChem and compare their results to GPT models, highlighting the increased complexity of ALAG compared to ASAG. Despite leveraging the benefits of a rubric-based approach and transfer learning from MNLI, the lower performance of LLMs on RiceChem underscores the significant difficulty posed by the ALAG task. With this work, we offer a fresh perspective on grading long, fact-based answers and introduce a new dataset to stimulate further research in this important area. Code: \url{https://github.com/luffycodes/Automated-Long-Answer-Grading}.

\end{abstract}

\section{Introduction}

The field of educational Natural Language Processing (NLP) has traditionally focused on short answer grading and open-ended essay grading. This paper explores an innovative, comparatively unexplored domain: Automated Long Answer Grading (ALAG). Unlike open-ended essays, which are assessed on traits such as coherence and originality \cite{nitinbook}, long answers are fact-based and require a different, more nuanced grading approach. Traditional ASAG methods \cite{burrows2015eras,bonthu2021automated,ramesh2022automated}, which utilize a 5-way classification system categorizing answers as `correct, partially correct, contradictory, irrelevant, or not in the domain', are not entirely suitable for grading long answers. This is due to the fact that long responses can simultaneously exhibit characteristics of multiple categories, rendering the 5-way ASAG classification formulation ineffective.

To enable a comprehensive study of ALAG, we curated a unique dataset, RiceChem, which consists of 1264 long answer responses from a college-level chemistry course.
\textit{RiceChem includes 1264 long answers, each graded against a subset of 27 rubric items, resulting in 8392 data points.}
The dataset is characterized by an average word count of 120, significantly higher than existing datasets such as SciEntsBank \cite{semeval2013}, Beetle \cite{semeval2013}, and Texas 2011 \cite{texas2011}, which have average word counts of 13, 10, and 18 respectively.
This stark difference in word count makes RiceChem an apt dataset for the exploration of ALAG.

\begin{figure*}[t]
  \centering
  \includegraphics[width=\textwidth]{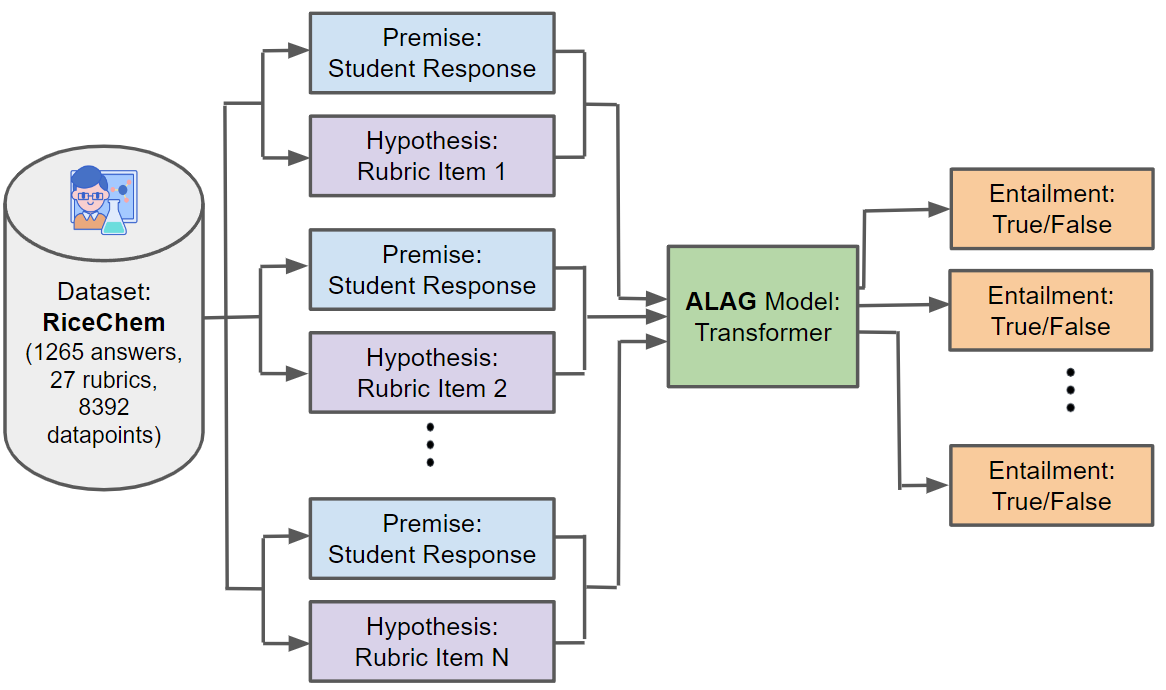}
  \caption{
  Schematic illustration of the Automated Long Answer Grading (ALAG) using the RiceChem dataset. The figure highlights our novel approach of formulating ALAG as a rubric entailment problem, where each student response (premise) is paired with a corresponding rubric item (hypothesis). These pairs are then processed by a fine-tuned ALAG-transformer model, which predicts whether the response entails the rubric item. The use of rubrics in RiceChem allows for a detailed, point-by-point evaluation, making the grading process interpretable by design.}
  \label{fig:flow}
\end{figure*}

In view of the shortcomings of traditional ASAG methods for ALAG, we redefine the problem as a rubric entailment task. In this novel formulation, each rubric item serves as a specific criterion that the student's answer should fulfill. We utilize natural language inference models to determine whether each rubric is entailed in the response, enabling a thorough and nuanced grading.

We set baselines for the ALAG task on the RiceChem dataset by fine-tuning encoder models such as BERT \cite{devlin2018bert}, RoBERTa \cite{roberta}, and BART \cite{lewis2019bart}. Our findings highlight the increased complexity and challenge of ALAG, even when the grading is facilitated by the rubric-based approach.
We demonstrate the importance of rubric-based formulation in ALAG, showcasing its superiority over traditional score-based approaches in capturing the nuances and multiple facets of student responses. Furthermore, we investigate the performance of models in cold start scenarios, providing valuable insights into the data efficiency and practical deployment considerations in educational settings.

Lastly, we benchmark state-of-the-art open-sourced Large Language Models (LLMs) \cite{phi2,team2024gemma,sonkar2023class,jiang2023mistral,sonkar2024code,bai2023qwen,sonkar2024pedagogical,young2024yi} on RiceChem and compare their results to GPT models \cite{gpt4sparkai}, highlighting the increased complexity of ALAG compared to ASAG. Despite leveraging the benefits of a rubric-based approach, the lower performance of LLMs on RiceChem, compared to ASAG's SciEntsBank, underscores the significant difficulty posed by the ALAG task.


\textbf{Contributions.} This work marks one of the first attempts, to our knowledge, at tackling Automated Long Answer Grading (ALAG) in the field of educational NLP. Our contributions are threefold. First, we provide a unique dataset, RiceChem, designed specifically for ALAG, to encourage further research in this crucial area of educational NLP. Second, we propose a new grading formulation, tailored to address the unique complexities of long answers. 
Finally, we present a comprehensive evaluation of state-of-the-art models, including LLMs, on the ALAG task, highlighting the challenges and opportunities for future research in this domain.

\section{Related Work: ALAG vs ASAG/AEG}
\textbf{ASAG Datasets.} Several ASAG datasets have been developed for different scales and domains, including SciEntsBank, Beetle, and Texas2011. These have been invaluable in advancing automated grading, however, their utility in the context of ALAG is limited. One of the reasons for this is the significant disparity in the average word count of responses in ASAG datasets (ranging from 10 to 18 words) compared to those in our RiceChem dataset (with an average word count of 120).

Furthermore, the grading systems in these ASAG datasets are based on a classification approach, where answers are categorized into `correct, partially correct, contradictory, irrelevant, or not in the domain' in the case of the 5-way categorization. 
\textit{While this system may be suitable for short answers, it fails to capture the complexity of long answers.}
In a long answer, multiple facets of an answer can simultaneously fall into different categories, leading to a situation where the existing classification system is insufficient.
For example, a long answer could contain elements that are correct, others that are partially correct, and still others that are irrelevant, leading to an overlap of categories that challenges the boundaries of the existing classification system. 
To rectify this shortcoming, we propose a more nuanced grading approach - a rubric-based grading system. 

\textbf{AEG Datasets.} 
Several datasets have been built for Automated Essay Grading (AEG), encompassing diverse scales and domains. Key examples include the Automated Student Assessment Prize (ASAP) dataset \cite{asap-aes}, which contains essays from 8 different prompts; the International Corpus of Learner English \cite{icle} consisting of essays from higher education students of varying English proficiency levels, evaluated on coherence, lexical richness, and grammatical accuracy; the Cambridge Learner Corpus \cite{openclc2017}, which comprises examination scripts from candidates taking Cambridge ESOL examinations, assessed on content, communicative quality, organization, and language use; and the TOEFL11 dataset \cite{toefl11}, which includes essays evaluated on development, organization, and language use, written by English learners for the Test of English as a Foreign Language (TOEFL) exam.

All these datasets, despite their diversity, have grading criteria notably different from those required for ALAG.
\textit{While essays emphasize attributes such as originality, coherence, and lexical richness, long answers necessitate a more fact-based grading approach, focusing on the accuracy and completeness of the information.}


\section{Dataset and Method}
\begin{figure*}[t]
    \centering
    \includegraphics[scale=0.60]{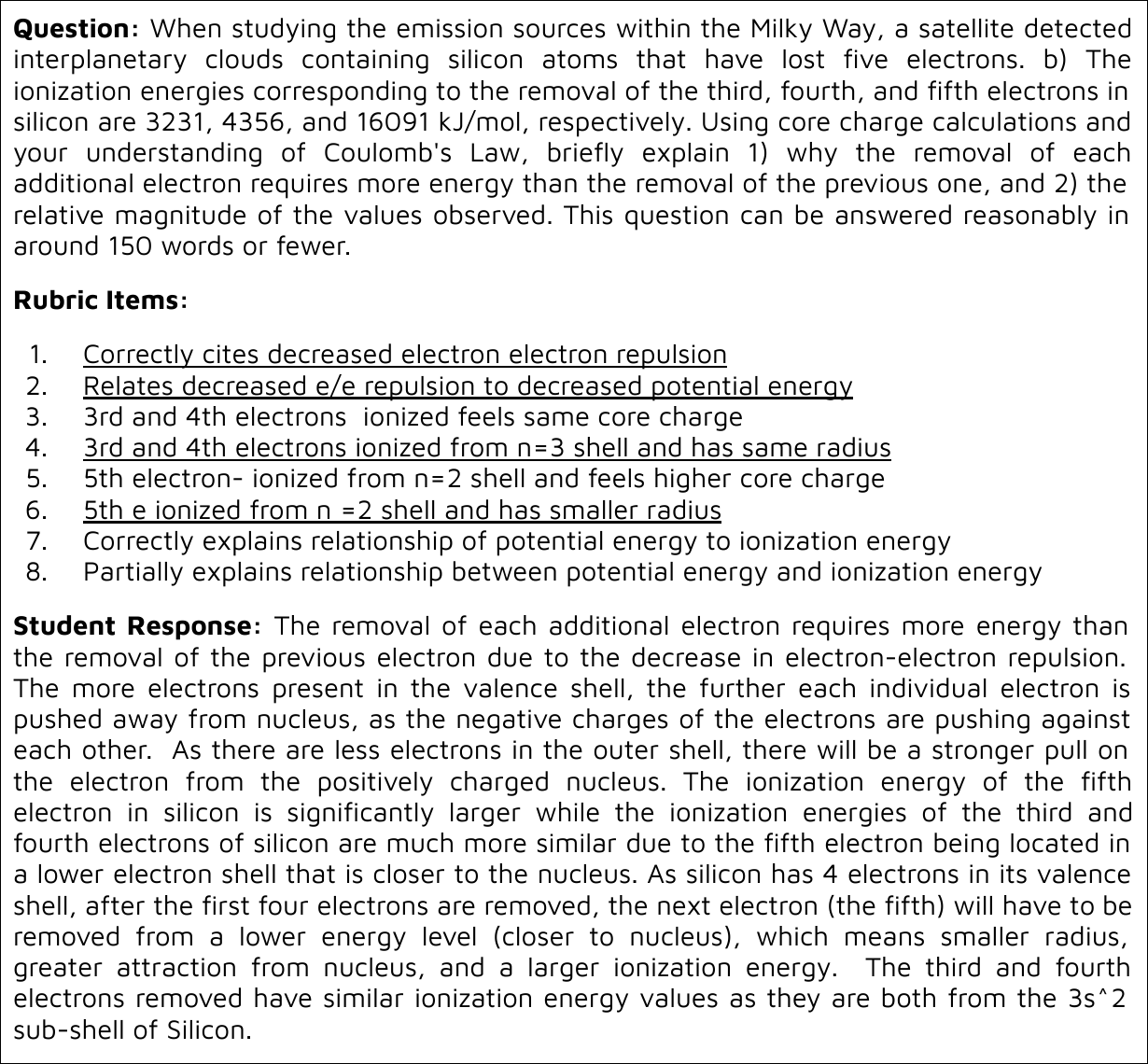}
    \caption{An example from our \dataset dataset showing a question, rubric items, and a student response.
    Underlined rubric items have been correctly answered by the student.}
    \label{fig:rubrichem_example}
    \vspace{-5mm}
\end{figure*}
In this section, we first introduce our unique \dataset dataset, followed by the problem formulation of the \model task. 
We also provide an overview of our proposed \model system, with all its components illustrated in figure~\ref{fig:flow}.

\subsection{\dataset Dataset}
To enable the exploration of the \model task, we have developed the \dataset dataset.
This dataset not only serves as a valuable resource for researchers working on ALAG, but also paves the way for the creation of more reliable and interpretable grading systems that can provide meaningful feedback to students through the use of rubrics.

\dataset features \textit{4 exam questions, 27 rubric items, and 1264 graded student responses} from a college-level chemistry course. Multiple teaching assistants graded the student responses against individual rubric items by assigning a TRUE or FALSE label. 
\textit{There are a total of 4880 TRUE labels and 3512 FALSE labels.}
Each rubric item holds a designated point value, and the final score for a response is determined by aggregating the scores of rubric items that are correctly answered.

\subsection{\longmodel (\modelns)}

Let an inference model $M: (P, H) \rightarrow {L}$ be give, which takes a premise $P$ and a hypothesis $H$ as inputs and predicts a label $L \in \{True, False\}$ indicating whether $P$ \textit{entails} $H$.
To formulate grading as an inference problem, a student response $R$ and a rubric item $I$ can be treated as the premise and hypothesis respectively such that $(R, I) \xrightarrow{M} L$.

Our \model approach implements the aforementioned formulation by training a  language model to predict the entailment of a rubric item from a student response.
The predictions can effectively pinpoint the correctly addressed rubric items in the student response, thus providing automated feedback.


\section{Experiments}

In this section, we provide a comprehensive overview of our experimental setup and results. We begin by detailing the training procedure for transformer language models on the RiceChem dataset and introducing the evaluation metrics used throughout our experiments. Next, we present the benchmarking results of various transformer models, including BERT, RoBERTa, and BART, on the ALAG task. We then highlight the importance of entailment-based and rubric-based formulation in ALAG and demonstrate its superiority over traditional score-based approaches. Furthermore, we investigate the performance of these models in cold start scenarios, where limited labeled data are available, and discuss the implications for practical deployment in educational settings.  Finally, we evaluate the performance of state-of-the-art open-sourced Large Language Models (LLMs) on RiceChem and compare their results to GPT models, showcasing the increased complexity of ALAG compared to ASAG.

\subsection{Experimental Setup}
To fine-tune the transformer models on \datasetns, we pre-process the data by employing an $80$-$10$-$10$ train-validation-test split. 
Specifically, for each question, we randomly select $80\%$ of student responses for training, $10\%$ for validation, and $10\%$ for testing to ensure three splits of responses are disjoint. 

We conduct the experiments using the Hugging Face transformers library \cite{wolf2020huggingfaces}.
The training process uses an NVIDIA A100-PCIE-40GB GPU.
During training, we use the AdamW optimizer \cite{loshchilov2019decoupled}, setting the initial learning rate to $2e^{-5}$.
Each update is performed with a mini-batch size of 16, and the model is trained for a maximum of 10 epochs.
The hyper-parameters $\alpha$ and $\beta$ are set to $0.9$ and $0.999$ respectively. 
After the training, we select the model with the highest validation F1 score as the best model for evaluations. 
For the experimental baselines, we employ a comprehensive set of evaluation metrics, including accuracy, precision, recall, and F1 score.
To ensure robustness, we report the averages and standard deviations of metrics across $5$ runs with $5$ different seeds.


\subsection{Benchmarking on Discriminative Models}
We assess the performance of state-of-the-art discriminative language models, such as BERT, RoBERTa, and BART, as benchmarks on the \dataset dataset.
In Table~\ref{tab:baselines}, we compare the results achieved by both base and large models. 
Notably, the large models outperform their base counterparts, demonstrating the advantages of employing more advanced models for this task. However, for the BERT model, it is not the case which can be attributed to the instability of fine-tuning it \cite{mosbach2021stability}.


\begin{table*}[t]
\centering
\normalsize
\caption{
Baseline performance of base and large transformer language models on the \dataset dataset for the \model task.
The figures in parentheses next to the model names represent the number of model parameters (in millions).
We report the mean and standard deviations across $5$ runs.
The large models exhibit superior performance compared to the base models (except the least performing BERT model), indicating that the task complexity requires more sophisticated models for \datasetns.
}
\resizebox{0.9\textwidth}{!}{
\begin{tabular}{l|l|l|l|l}
  \hline
  \textbf{Model} & \textbf{Accuracy} & \textbf{Precision} & \textbf{Recall} & \textbf{F1} \\
  \hline
  RoBERTa-base (125M) & 83.0 (0.7) & 0.830 (0.020) & 0.883 (0.020) & 0.856 (0.004) \\ 
  RoBERTa-large (355M) & \textbf{84.1} (0.9) & \textbf{0.840} (0.009) & 0.891 (0.015) & \textbf{0.864} (0.008) \\
  BART-base (140M) & 83.6 (1.2)& 0.832 (0.021)& 0.892   (0.014)& 0.861 (0.009)\\
  BART-large (406M) & 83.9 (0.9)& 0.833 (0.009) & \textbf{0.897} (0.007) & \textbf{0.864} (0.007)\\
  BERT-base (110M) & 82.8 (1.1) & 0.828 (0.019) & 0.882 (0.018) & 0.854 (0.008) \\
  BERT-large (340M) & 82.5 (0.5) & 0.825 (0.009)& 0.879 (0.015)& 0.851 (0.005) \\
  \hline
\end{tabular}
}

\label{tab:baselines} 
\end{table*}
\begin{table}[t]
\centering
\caption{
Performance comparison of large models and their MNLI fine-tuned versions on the RiceChem dataset. RoBERTa and BART show accuracy increases by $3.2\%$ and $1.8\%$, and F1 score increases by $2.8\%$ and $1.4\%$. This highlights the benefit of formulating \model as an entailment problem, enabling the use of the large MNLI dataset for performance enhancement.}
\normalsize
\resizebox{0.55\linewidth}{!}{
\begin{tabular}{l|l|l}
  \hline
  \textbf{Model} & \textbf{Accuracy} & \textbf{F1} \\
  \hline
  RoBERTa-large & 84.1 & 0.864 \\
  RoBERTa-large-mnli & \textbf{86.8} & \textbf{0.888} \\  
  BART-large & 83.9 & 0.864 \\
  BART-large-mnli & 85.4 & 0.876 \\  
  \hline
\end{tabular}
}
\label{tab:baselines_mnli}
\end{table}


\subsection{The Value of Entailment Formulation in ALAG}
In Table~\ref{tab:baselines_mnli}, we compare the performance of language models and their MNLI fine-tuned counterparts on the RiceChem dataset. The results demonstrate a significant improvement in both accuracy and F1 score when the models are fine-tuned on the MNLI (Multi-Genre Natural Language Inference Corpus) dataset \cite{mnli}, highlighting the value of formulating ALAG as an entailment problem.

By framing ALAG as an entailment task, we enable the use of the MNLI dataset, which contains a diverse set of premise-hypothesis pairs covering a wide range of topics and linguistic variations. The MNLI dataset, with its 4 million examples, provides a rich source of linguistic knowledge and reasoning capabilities that can be effectively transferred to the ALAG task.

The entailment formulation allows us to leverage the models pre-trained on the MNLI dataset, which have already acquired a strong understanding of the entailment relationship between premises and hypotheses. By fine-tuning these models on the RiceChem dataset, we can efficiently transfer the learned knowledge and adapt it to the specific domain of long answer grading.

The performance gains observed in Table~\ref{tab:baselines_mnli} underscore the effectiveness of this transfer learning approach. The RoBERTa model, when fine-tuned on MNLI, exhibits a 3.2\% increase in accuracy and a 2.8\% increase in F1 score compared to its vanilla counterpart. Similarly, the BART model shows a 1.8\% increase in accuracy and a 1.4\% increase in F1 score. These improvements demonstrate the successful transfer of knowledge from the MNLI dataset to the ALAG task, enabled by the entailment formulation.

The entailment formulation not only facilitates the use of large-scale datasets like MNLI but also provides a more natural and interpretable approach to ALAG. By aligning the grading process with the task of determining the entailment relationship between the student response and the rubric item, we create a more intuitive and explainable framework for assessing student responses.


\begin{figure*}[t!]
  \centering
  \begin{subfigure}[b]{0.495\textwidth}
    \centering
    \includegraphics[width=\textwidth]{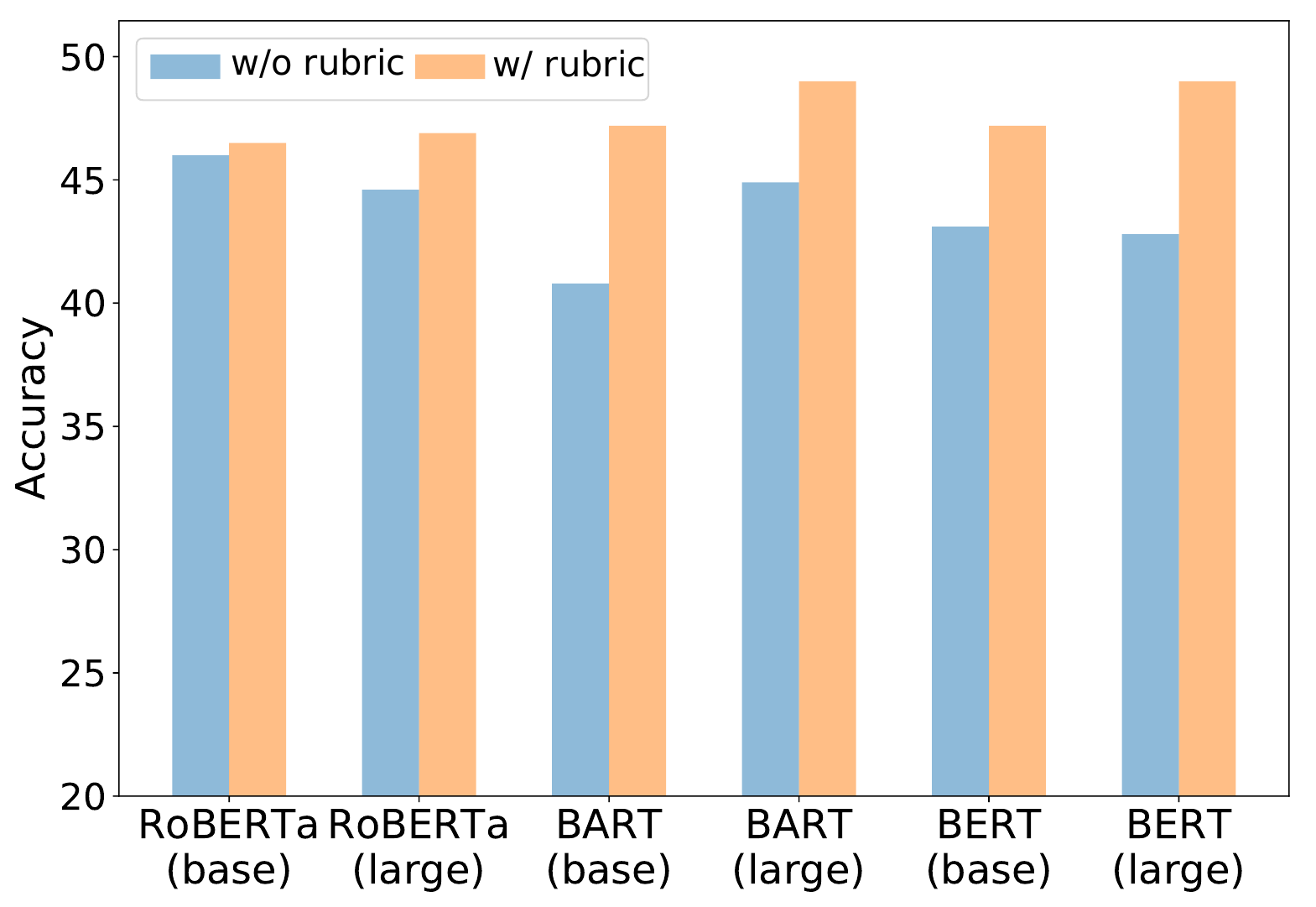}
  \end{subfigure}
  \hfill
  \begin{subfigure}[b]{0.495\textwidth}
    \centering
    \includegraphics[width=\textwidth]{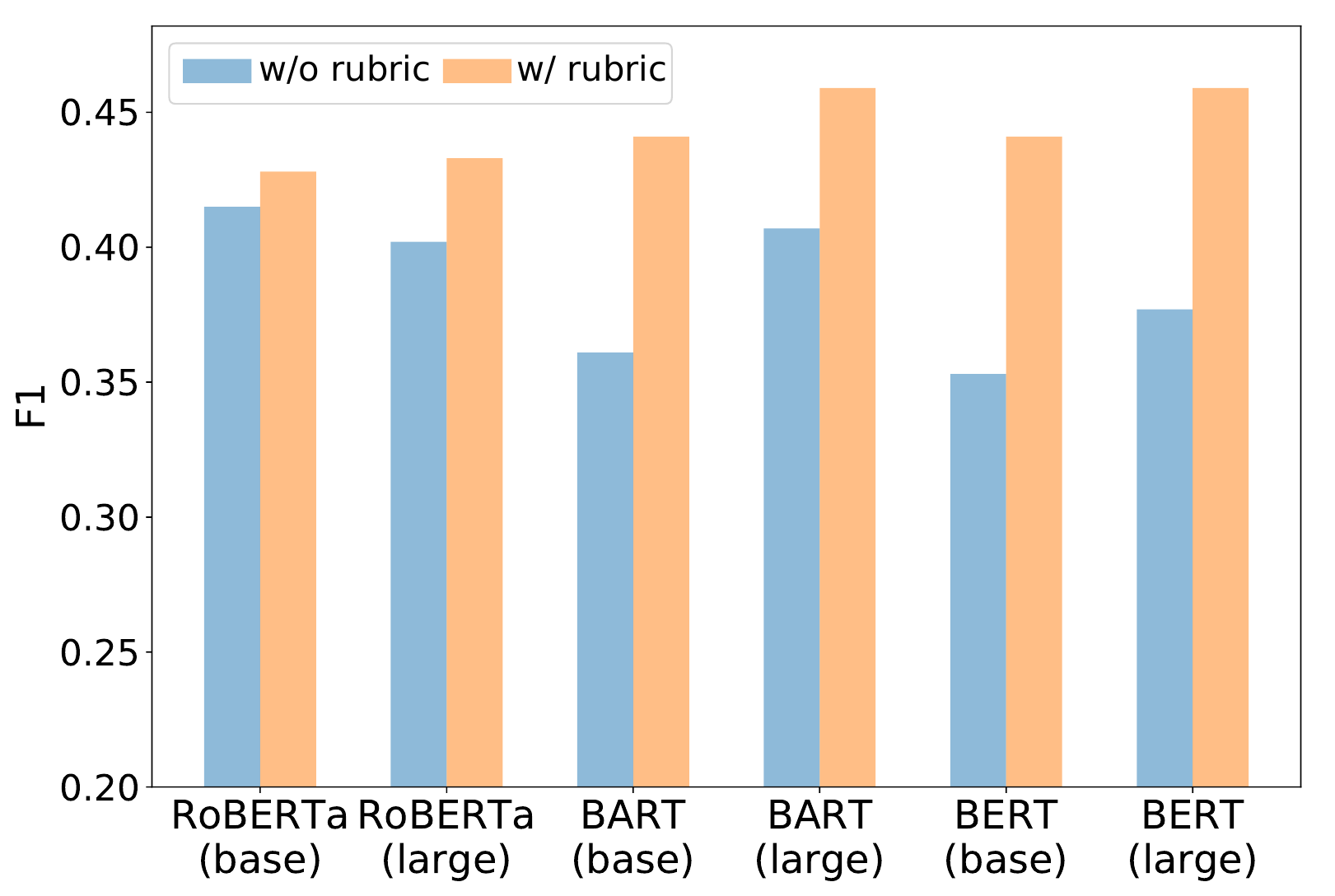}
  \end{subfigure}
  \captionsetup{skip=0pt}
  \caption{Comparisons between the traditional score based grading approach and rubric-based \model approach on the \dataset dataset.
  Rubric-based \model offers an average increase of $9.2\%$ in accuracy and an average increase of $15.4\%$ in F1 score, proving that breaking down grading into smaller rubric items helps models focus on smaller parts of the task instead of doing the entire task altogether. The improvement is evident across all models regardless of their number of parameters.}
  \label{fig:comparison_acc}
\end{figure*}
\subsection{The Importance of Rubric-based Formulation in ALAG}
The use of rubrics in automated grading has been shown to lead to performance improvements in Automated Short Answer Grading (ASAG) \cite{rubricasag} and Automated Essay Grading (AEG) \cite{rubricessay}. Our experiments confirm that this holds true for Automated Long Answer Grading (ALAG) as well, with the rubric-based approach demonstrating an average improvement of 9.2\% in accuracy and 15.4\% in F1 score compared to the traditional score-based approach (Figure~\ref{fig:comparison_acc}).

The use of rubrics in automated grading has been shown to lead to performance improvements in previous works on ASAG \cite{rubricasag} and AEG \cite{rubricessay}. Our experiments confirm that this holds true for ALAG as well. However, the importance of rubric-based formulation in ALAG is even more pronounced due to the complexity and multifaceted nature of long answers.

To illustrate this, we compare our rubric-based ALAG approach to a traditional score-based approach. In the score-based approach, we pre-process the RiceChem dataset by structuring the data into sentences (student responses) and labels (scores), with the language model aiming to predict an integer score from 0 to 8. On the other hand, our rubric-based ALAG formulation breaks down the grading process into smaller, more manageable components, allowing the model to focus on specific aspects of the answer as defined by the rubric items.

Figure~\ref{fig:comparison_acc} demonstrates the superiority of the rubric-based approach, with an average improvement of 9.2\% in accuracy and 15.4\% in F1 score compared to the traditional score-based approach. This significant performance gain highlights the importance of leveraging rubrics in ALAG. By decomposing the complex task of grading long answers into smaller, well-defined rubric items, the model can more effectively capture the nuances and multiple facets of the student responses.

It is worth noting that creating high-quality rubrics is a challenging task that requires careful consideration and effort. However, this effort needs to be invested only once, and the benefits of a well-designed rubric can be reaped repeatedly in the automated grading process. The rubric serves as a comprehensive framework that guides the model in assessing the key aspects of the answer, ensuring a more accurate and reliable grading outcome.

Moreover, the use of rubrics in ALAG not only improves the model's performance but also enhances the interpretability and transparency of the grading process. By aligning the model's predictions with specific rubric items, educators and students can gain a clearer understanding of the strengths and weaknesses of the answers, facilitating targeted feedback and improvement.

\subsection{Benchmarking on Cold Start Scenarios}
\begin{table}[tb!]
\caption{Performance of RoBERTa-Large-MNLI on unseen questions. The model is trained on three questions and evaluated on the remaining question, demonstrating its ability to generalize to new questions without prior training data.}
\centering
\begin{tabular}{l|l|l|l|l}
\hline
\textbf{Unseen} & \textbf{Accuracy} & \textbf{Precision} & \textbf{Recall} & \textbf{F1} \\
\hline
Q1 & 65.9 & 0.703 & 0.717 & 0.705 \\
Q2 & 68.7 & 0.704 & 0.584 & 0.629 \\
Q3 & 66.7 & 0.649 & 0.644 & 0.633 \\
Q4 & 60.6 & 0.892 & 0.611 & 0.717 \\
\hline
\end{tabular}
\label{tab:unseen_data}
\end{table}
In educational settings, it is common to encounter situations where limited labeled data is available for training automated grading models, especially when dealing with new courses, subjects, or question types. Therefore, it is crucial to assess the performance of models in cold start settings and understand how their performance evolves as more training data becomes available. The analysis in this section provides valuable insights into the data efficiency of the models and helps determine the minimum amount of labeled data required to achieve satisfactory grading results.

We begin by evaluating the performance of the RoBERTa-Large-MNLI model on unseen questions, simulating a scenario where the model is fine-tuned on some questions but is applied to grade responses for a completely new question without any prior training data. For this type of investigation, we trained the model on three questions in the dataset, and used the remaining unseen question for testing. As shown in Table~\ref{tab:unseen_data}, the model demonstrates a reasonable level of generalization, with accuracy ranging from 60.6\% to 68.7\% and F1 scores ranging from 0.629 to 0.717 across the four questions. This suggests that the model, fine-tuned on similar types of questions, has acquired some level of transferable knowledge that enables it to handle unseen questions to a certain extent, which can be valuable in educational settings where labeled data for new questions may be scarce.

\begin{figure}[t!]
    \centering
    \begin{subfigure}{0.49\textwidth}
        \centering
        \includegraphics[width=\textwidth]{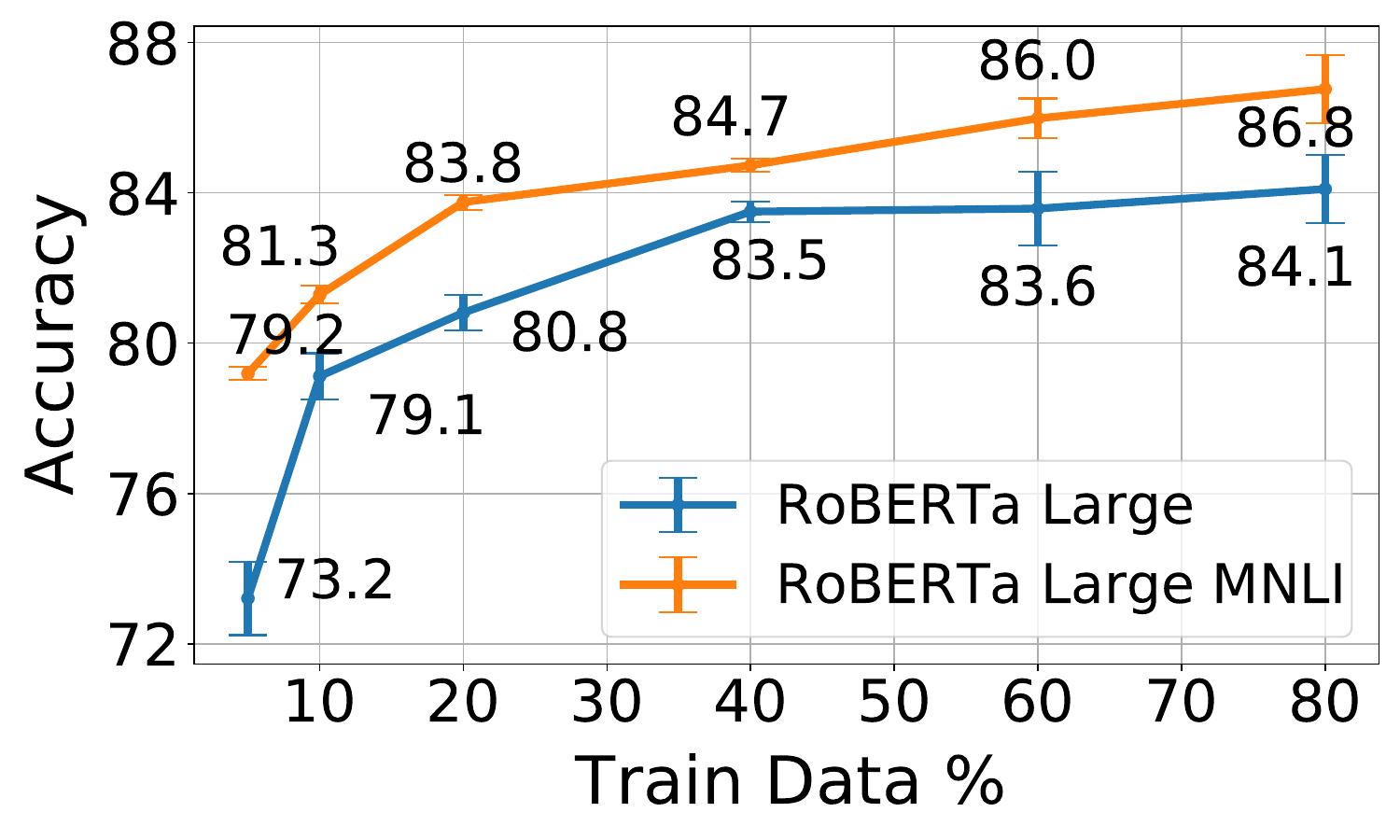}
        \caption{Accuracy of RoBERTa models}
        \label{fig:exp1_acc}
    \end{subfigure}\hfill
    \begin{subfigure}{0.49\textwidth}
        \centering
        \includegraphics[width=\textwidth]{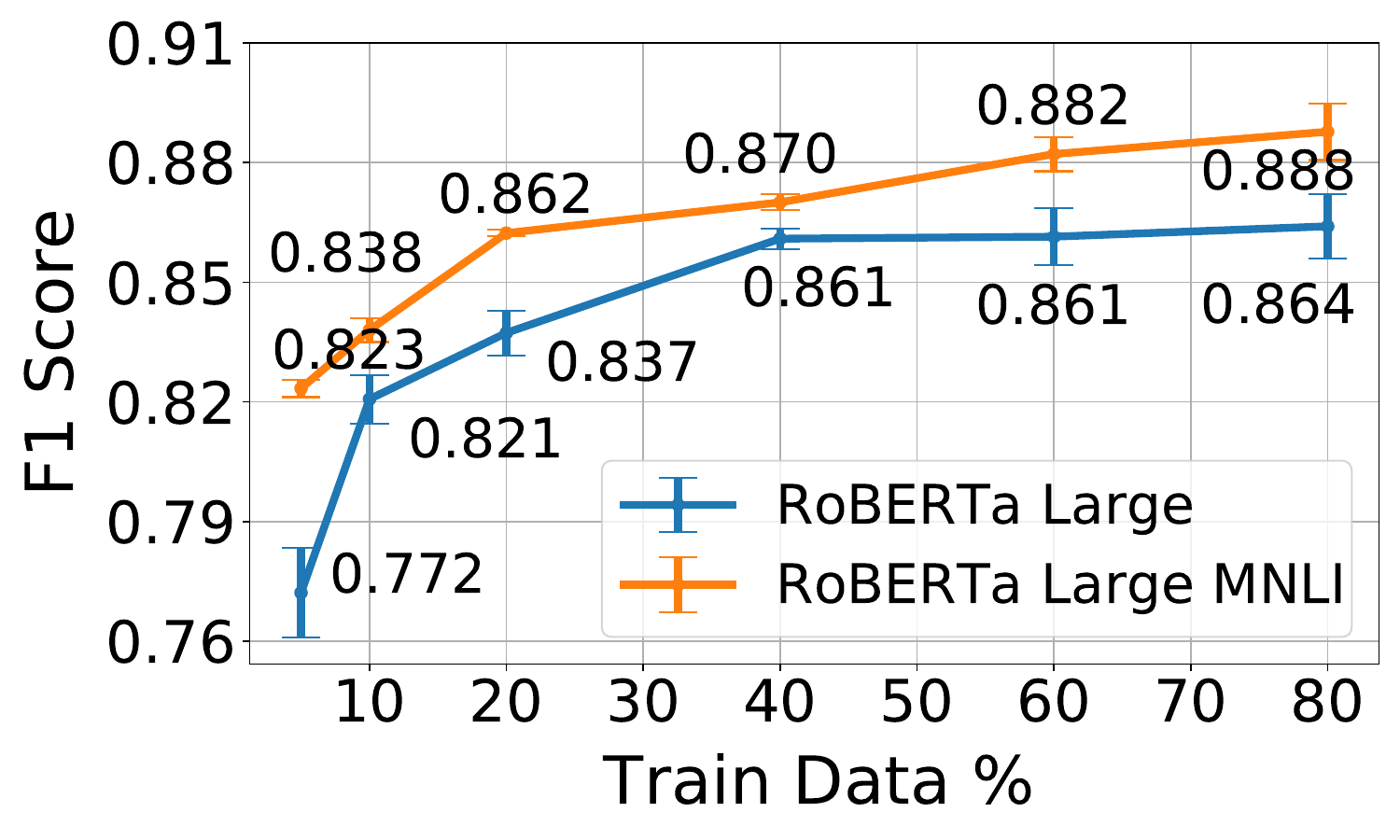}
        \caption{F1 Score of RoBERTa models}
        \label{fig:exp1_f1}
    \end{subfigure}
    
    \caption{Performance of RoBERTa-Large and RoBERTa-Large-MNLI models with varying amounts of training data, ranging from 5\% to 80\%. The models show consistent improvement in accuracy and F1 score as more labeled data becomes available, with diminishing returns after 40\% for RoBERTa-Large and 20\% for RoBERTa-Large-MNLI.}
    \label{fig:different_training_splits}
\end{figure}

Next, we investigate the performance of the RoBERTa-Large model and its MNLI fine-tuned version as the amount of training data increases from 5\% to 80\%. Figure~\ref{fig:different_training_splits} illustrates the trends in accuracy and F1 score for both models. As expected, there is a consistent improvement in performance as more training data becomes available. For RoBERTa Large, the accuracy increases from 73.2\% to 84.1\%, and the F1 score rises from 0.772 to 0.864. Similarly, for the MNLI fine-tuned version, the accuracy improves from 79.2\% to 86.8\%, and the F1 score increases from 0.823 to 0.888.

Interestingly, the performance gains exhibit diminishing returns, particularly after 40\% of the training data for RoBERTa-Large and 20\% for RoBERTa-Large-MNLI. This observation suggests that the models can achieve competitive grading results even with a relatively small amount of labeled data, and the benefits of additional data become less pronounced beyond a certain point. Moreover, the standard deviations for both accuracy and F1 score remain within 1.12\% across different seeds, indicating the reliability and consistency of the models' performance.

These findings have important implications for practical deployment scenarios in educational institutions. The analysis provides guidance on the trade-off between labeling effort and performance gains, allowing educators and administrators to make informed decisions based on their specific requirements and resource constraints when implementing automated grading systems.

\subsection{Benchmarking on Large Language Models}
\label{app:generative}
\begin{table}[tb!]
\centering
\caption{Zero-shot performance comparison of Large Language Models (LLMs) on the RiceChem dataset for the Automated Long Answer Grading (ALAG) task. The results highlight the complexity of ALAG, with even the best-performing model, GPT-4, achieving an accuracy of 70.9\% and an F1 score of 0.689. The performance gap between LLMs on ALAG and their performance on Automated Short Answer Grading (ASAG) tasks emphasizes the unique challenges posed by grading long, fact-based answers.}
\begin{tabular}{l|l|l|l|l}
\hline
\textbf{LLM}         & \textbf{Accuracy} & \textbf{Precision} & \textbf{Recall} & \textbf{F1} \\
\hline
Phi 2 \cite{phi2}                   & 58.65 & 64.15 & 9.16  & 16.04 \\
Gemma 7B 1.1 IT  \cite{team2024gemma}          & 60.05 & 58.71 & 24.53 & 34.6  \\
Mistral 7B Instruct-v0.2 \cite{jiang2023mistral}  & 61.9  & 60.59 & 33.15 & 42.86 \\
OLMo 7B Instruct     \cite{groeneveld2024olmo}      & 60.39 & 57.5  & 31.0  & 40.28 \\
Zephyr 7B Beta    \cite{tunstall2023zephyr}         & 57.03 & 50.6  & 11.32 & 18.5  \\
Vicuna 13B v1.5   \cite{vicuna2023}         & 59.58 & \textbf{81.08} & 8.09  & 14.71 \\
Qwen1.5 14B Chat   \cite{bai2023qwen}        & 58.65 & 63.64 & 9.43  & 16.43 \\
Qwen1.5 32B Chat   \cite{bai2023qwen}        & 62.37 & 60.44 & 36.66 & 45.64 \\
Yi 34B Chat    \cite{young2024yi}            & 58.19 & 53.22 & 24.53 & 33.5  \\
GPT3.5      \cite{gpt3human}               & 67.1  & 59.1  & 45.7  & 51.6  \\
GPT4       \cite{gpt4asag}                & \textbf{70.9} & 63.9  & \textbf{74.8}  & \textbf{68.9}  \\
\hline
\end{tabular}
\label{tab:baselines_llm}
\end{table}
The rapid advancements in LLMs have led to significant improvements in various natural language processing tasks. To assess the potential of these models in the context of ALAG, we evaluate the zero-shot performance of several LLMs on the RiceChem dataset, as presented in Table~\ref{tab:baselines_llm}. For GPT models \cite{gpt4sparkai}, we used the following hyperparameters: a temperature of 1.0, a frequency penalty of 0, and a presence penalty of 0. For open-sourced LLMs \cite{phi2,team2024gemma,jiang2023mistral,groeneveld2024olmo,tunstall2023zephyr,vicuna2023,bai2023qwen,young2024yi}, we considered the log probabilities of the `True' and `False' tokens to determine if the student answered the rubric item correctly or not. By comparing these log probabilities, we obtained the model's predicted label.

Despite the impressive success of LLMs in many domains, the RiceChem dataset proves to be a formidable challenge. The best-performing model, GPT-4, achieves an accuracy of 70.9\% and an F1 score of 0.689, highlighting the complexity of the ALAG task. This performance is particularly striking when compared to the results of GPT models on ASAG tasks.

Previous research has shown that GPT models can achieve an F1 score of 0.74 on the SciEntsBank dataset for ASAG without the use of rubrics  \cite{gpt4asag}. In contrast, GPT-4 obtains a lower F1 score of 0.69 on RiceChem, despite the significant beneficial impact of rubrics typically observed in ASAG \cite{rubricasag} and AEG \cite{rubricessay} tasks. This discrepancy underscores the increased difficulty of ALAG compared to ASAG.

It is worth noting that the actual difference in complexity between ASAG and ALAG may be even more substantial than the five-point difference in F1 scores suggests. The use of rubrics in RiceChem provides a structured framework for grading, which is expected to enhance model performance. However, even with this advantage, GPT-4 still struggles to match its performance on ASAG tasks without rubrics.

The results in Table~\ref{tab:baselines_llm} also reveal the varying performance of different LLMs on the RiceChem dataset. While GPT-4 and GPT-3.5 stand out as the top performers, other models such as Qwen1.5 32B Chat \cite{bai2023qwen} and Mistral \cite{jiang2023mistral} show promising results, with F1 scores of 0.456 and 0.429, respectively. These findings indicate that the choice of LLM architecture and training methodology can have a significant impact on the model's ability to handle the complexities of ALAG.

In summary, the benchmarking of LLMs on the RiceChem dataset highlights the unique challenges posed by the ALAG task. The performance gap between GPT models on ASAG and ALAG tasks, even with the benefit of rubrics, emphasizes the need for further research and development of specialized models and techniques to effectively tackle the complexities of grading long, fact-based answers. As LLMs continue to evolve, it will be crucial to explore their potential in the context of ALAG and develop strategies to harness their capabilities for improving automated grading systems in educational settings.

\section{Conclusion}
In this paper, we introduce the novel task of Automated Long Answer Grading (ALAG) and present the RiceChem dataset, specifically designed to facilitate research in this domain. Our rubric-based formulation of ALAG provides a nuanced and pedagogically sound approach to evaluating long answers, offering a more comprehensive assessment compared to traditional ASAG methods. Through extensive experiments, we demonstrate the importance of rubric-based formulation, the value of entailment formulation, and the challenges posed by cold start scenarios. Furthermore, our benchmarking of state-of-the-art models, including LLMs, confirms that ALAG poses a significantly greater challenge compared to ASAG. We believe this work will stimulate and inspire further research in this crucial area of educational NLP, contributing to the development of advanced models capable of handling the complexities and intricacies of the ALAG task.

\section*{Acknowledgments}
This work was supported by NSF grants 1842378, ONR grant N0014-20-1-2534, AFOSR grant FA9550-22-1-0060, and a Vannevar Bush Faculty Fellowship, ONR grant N00014-18-1-2047.

\bibliographystyle{unsrt}  
\bibliography{custom}

\begin{thebibliography}{10}

\bibitem{nitinbook}
Beata~Beigman Klebanov and Nitin Madnani.
\newblock {\em Automated Essay Scoring}.
\newblock Springer Nature, 2022.

\bibitem{burrows2015eras}
Steven Burrows, Iryna Gurevych, and Benno Stein.
\newblock The eras and trends of automatic short answer grading.
\newblock {\em International Journal of Artificial Intelligence in Education}, 25:60--117, 2015.

\bibitem{bonthu2021automated}
Sridevi Bonthu, S~Rama~Sree, and MHM Krishna~Prasad.
\newblock Automated short answer grading using deep learning: A survey.
\newblock In {\em Machine Learning and Knowledge Extraction: 5th IFIP}, pages 61--78. Springer, 2021.

\bibitem{ramesh2022automated}
Dadi Ramesh and Suresh~Kumar Sanampudi.
\newblock An automated essay scoring systems: a systematic literature review.
\newblock {\em Artificial Intelligence Review}, 55(3):2495--2527, 2022.

\bibitem{semeval2013}
Myroslava Dzikovska, Rodney Nielsen, Chris Brew, Claudia Leacock, Danilo Giampiccolo, Luisa Bentivogli, Peter Clark, Ido Dagan, and Hoa~Trang Dang.
\newblock Semeval-2013 task 7: The joint student response analysis and 8th recognizing textual entailment challenge.
\newblock In {\em Proceedings of the Second Joint Conference on Lexical and Computational Semantics}, volume~2, pages 263--274, 2013.

\bibitem{texas2011}
Michael Mohler, Razvan Bunescu, and Rada Mihalcea.
\newblock Learning to grade short answer questions using semantic similarity measures and dependency graph alignments.
\newblock In {\em Proceedings of the 49th Annual Meeting of the Association for Computational Linguistics: Human Language Technologies}, pages 752--762, 2011.

\bibitem{devlin2018bert}
Jacob Devlin, Ming-Wei Chang, Kenton Lee, and Kristina Toutanova.
\newblock Bert: Pre-training of deep bidirectional transformers for language understanding.
\newblock {\em arXiv preprint arXiv:1810.04805}, 2018.

\bibitem{roberta}
Yinhan Liu, Myle Ott, Naman Goyal, Jingfei Du, Mandar Joshi, Danqi Chen, Omer Levy, Mike Lewis, Luke Zettlemoyer, and Veselin Stoyanov.
\newblock Roberta: {A} robustly optimized {BERT} pretraining approach.
\newblock {\em CoRR}, abs/1907.11692, 2019.

\bibitem{lewis2019bart}
Mike Lewis, Yinhan Liu, Naman Goyal, Marjan Ghazvininejad, Abdelrahman Mohamed, Omer Levy, Ves Stoyanov, and Luke Zettlemoyer.
\newblock Bart: Denoising sequence-to-sequence pre-training for natural language generation, translation, and comprehension, 2019.

\bibitem{phi2}
Yuanzhi Li, S{\'e}bastien Bubeck, Ronen Eldan, Allie Del~Giorno, Suriya Gunasekar, and Yin~Tat Lee.
\newblock Textbooks are all you need ii: phi-1.5 technical report.
\newblock {\em arXiv preprint arXiv:2309.05463}, 2023.

\bibitem{team2024gemma}
Gemma Team, Thomas Mesnard, Cassidy Hardin, Robert Dadashi, Surya Bhupatiraju, Shreya Pathak, Laurent Sifre, Morgane Rivi{\`e}re, Mihir~Sanjay Kale, Juliette Love, et~al.
\newblock Gemma: Open models based on gemini research and technology.
\newblock {\em arXiv preprint arXiv:2403.08295}, 2024.

\bibitem{sonkar2023class}
Shashank Sonkar, Naiming Liu, Debshila Mallick, and Richard Baraniuk.
\newblock Class: A design framework for building intelligent tutoring systems based on learning science principles.
\newblock In {\em Findings of the Association for Computational Linguistics: EMNLP 2023}, pages 1941--1961, 2023.

\bibitem{jiang2023mistral}
Albert~Q Jiang, Alexandre Sablayrolles, Arthur Mensch, Chris Bamford, Devendra~Singh Chaplot, Diego de~las Casas, Florian Bressand, Gianna Lengyel, Guillaume Lample, Lucile Saulnier, et~al.
\newblock Mistral 7b.
\newblock {\em arXiv preprint arXiv:2310.06825}, 2023.

\bibitem{sonkar2024code}
Shashank Sonkar, Xinghe Chen, Myco Le, Naiming Liu, Debshila Basu~Mallick, and Richard Baraniuk.
\newblock Code soliloquies for accurate calculations in large language models.
\newblock In {\em Proceedings of the 14th Learning Analytics and Knowledge Conference}, pages 828--835, 2024.

\bibitem{bai2023qwen}
Jinze Bai, Shuai Bai, Yunfei Chu, Zeyu Cui, Kai Dang, Xiaodong Deng, Yang Fan, Wenbin Ge, Yu~Han, Fei Huang, et~al.
\newblock Qwen technical report.
\newblock {\em arXiv preprint arXiv:2309.16609}, 2023.

\bibitem{sonkar2024pedagogical}
Shashank Sonkar, Kangqi Ni, Sapana Chaudhary, and Richard~G Baraniuk.
\newblock Pedagogical alignment of large language models.
\newblock {\em arXiv preprint arXiv:2402.05000}, 2024.

\bibitem{young2024yi}
Alex Young, Bei Chen, Chao Li, Chengen Huang, Ge~Zhang, Guanwei Zhang, Heng Li, Jiangcheng Zhu, Jianqun Chen, Jing Chang, et~al.
\newblock Yi: Open foundation models by 01. ai.
\newblock {\em arXiv preprint arXiv:2403.04652}, 2024.

\bibitem{gpt4sparkai}
S{\'e}bastien Bubeck, Varun Chandrasekaran, Ronen Eldan, Johannes Gehrke, Eric Horvitz, Ece Kamar, Peter Lee, Yin~Tat Lee, Yuanzhi Li, Scott Lundberg, et~al.
\newblock {Sparks of Artificial General Intelligence: Early experiments with GPT-4}.
\newblock {\em arXiv preprint arXiv:2303.12712}, 2023.

\bibitem{asap-aes}
Ben Hamner, Jaison Morgan, Mark~Shermis Lynnvandev, and Tom~Vander Ark.
\newblock The hewlett foundation: Automated essay scoring, 2012.

\bibitem{icle}
Sylviane Granger, Estelle Dagneaux, Fanny Meunier, Magali Paquot, et~al.
\newblock {\em International corpus of learner English}, volume~2.
\newblock Presses universitaires de Louvain Louvain-la-Neuve, 2009.

\bibitem{openclc2017}
Diane Nicholls.
\newblock The cambridge learner corpus: Error coding and analysis for lexicography and elt.
\newblock In {\em Proceedings of the Corpus Linguistics 2003 conference}, volume~16, pages 572--581. Cambridge University Press Cambridge, 2003.

\bibitem{toefl11}
Daniel Blanchard, Joel Tetreault, Derrick Higgins, Aoife Cahill, and Martin Chodorow.
\newblock Toefl11: A corpus of non-native english.
\newblock {\em ETS Research Report Series}, 2013(2):i--15, 2013.

\bibitem{wolf2020huggingfaces}
Thomas Wolf, Lysandre Debut, Victor Sanh, Julien Chaumond, Clement Delangue, Anthony Moi, Pierric Cistac, Tim Rault, Rémi Louf, Morgan Funtowicz, Joe Davison, Sam Shleifer, Patrick von Platen, Clara Ma, Yacine Jernite, Julien Plu, Canwen Xu, Teven~Le Scao, Sylvain Gugger, Mariama Drame, Quentin Lhoest, and Alexander~M. Rush.
\newblock Huggingface's transformers: State-of-the-art natural language processing, 2020.

\bibitem{loshchilov2019decoupled}
Ilya Loshchilov and Frank Hutter.
\newblock Decoupled weight decay regularization, 2019.

\bibitem{mosbach2021stability}
Marius Mosbach, Maksym Andriushchenko, and Dietrich Klakow.
\newblock On the stability of fine-tuning bert: Misconceptions, explanations, and strong baselines, 2021.

\bibitem{mnli}
Adina Williams, Nikita Nangia, and Samuel~R. Bowman.
\newblock A broad-coverage challenge corpus for sentence understanding through inference.
\newblock {\em CoRR}, abs/1704.05426, 2017.

\bibitem{rubricasag}
Smit Marvaniya, Swarnadeep Saha, Tejas~I Dhamecha, Peter Foltz, Renuka Sindhgatta, and Bikram Sengupta.
\newblock Creating scoring rubric from representative student answers for improved short answer grading.
\newblock In {\em Proceedings of the 27th ACM International Conference on Information and Knowledge Management}, pages 993--1002, 2018.

\bibitem{rubricessay}
Rahul Kumar, Sandeep Mathias, Sriparna Saha, and Pushpak Bhattacharyya.
\newblock Many hands make light work: Using essay traits to automatically score essays.
\newblock {\em arXiv preprint arXiv:2102.00781}, 2021.

\bibitem{groeneveld2024olmo}
Dirk Groeneveld, Iz~Beltagy, Pete Walsh, Akshita Bhagia, Rodney Kinney, Oyvind Tafjord, Ananya~Harsh Jha, Hamish Ivison, Ian Magnusson, Yizhong Wang, et~al.
\newblock Olmo: Accelerating the science of language models.
\newblock {\em arXiv preprint arXiv:2402.00838}, 2024.

\bibitem{tunstall2023zephyr}
Lewis Tunstall, Edward Beeching, Nathan Lambert, Nazneen Rajani, Kashif Rasul, Younes Belkada, Shengyi Huang, Leandro von Werra, Cl{\'e}mentine Fourrier, Nathan Habib, et~al.
\newblock Zephyr: Direct distillation of lm alignment.
\newblock {\em arXiv preprint arXiv:2310.16944}, 2023.

\bibitem{vicuna2023}
Wei-Lin Chiang, Zhuohan Li, Zi~Lin, Ying Sheng, Zhanghao Wu, Hao Zhang, Lianmin Zheng, Siyuan Zhuang, Yonghao Zhuang, Joseph~E. Gonzalez, Ion Stoica, and Eric~P. Xing.
\newblock Vicuna: An open-source chatbot impressing gpt-4 with 90\%* chatgpt quality, March 2023.

\bibitem{gpt3human}
Long Ouyang, Jeffrey Wu, Xu~Jiang, Diogo Almeida, Carroll Wainwright, Pamela Mishkin, Chong Zhang, Sandhini Agarwal, Katarina Slama, Alex Ray, et~al.
\newblock Training language models to follow instructions with human feedback.
\newblock {\em Advances in Neural Information Processing Systems}, 35:27730--27744, 2022.

\bibitem{gpt4asag}
Gerd Kortemeyer.
\newblock Performance of the pre-trained large language model gpt-4 on automated short answer grading.
\newblock {\em arXiv preprint arXiv:2309.09338}, 2023.

\end{thebibliography}

\end{document}